# DCT BASED TEXTURE CLASSIFICATION USING SOFT COMPUTING APPROACH

**GOLAM SORWAR***

*School of Multimedia and Information Technology*
*Southern Cross University*
*Coffs Harbour, NSW 2457, Australia*
*gsorwar@scu.edu.au*

**AJITH ABRAHAM**

*Computer Science Department*
*Oklahoma State University (Tulsa)*
*Tulsa, OK 74106, USA*
*ajith.abraham@ieee.org*

# DCT BASED TEXTURE CLASSIFICATION USING SOFT COMPUTING APPROACH

*ABSTRACT*

Classification of texture pattern is one of the most important problems in pattern recognition. In this paper, we present a classification method based on the Discrete Cosine Transform (DCT) coefficients of texture image. As DCT works on gray level image, the color scheme of each image is transformed into gray levels. For classifying the images using DCT we used two popular soft computing techniques namely neurocomputing and neuro-fuzzy computing. We used a feedforward neural network trained using the backpropagation learning and an evolving fuzzy neural network to classify the textures. The soft computing models were trained using 80% of the texture data and remaining was used for testing and validation purposes. A performance comparison was made among the soft computing models for the texture classification problem. We also analyzed the effects of prolonged training of neural networks. It is observed that the proposed neuro-fuzzy model performed better than neural network.

**Keywords:** *Texture classification, DCT, Neurocomputing, Neuro-Fuzzy Computing, soft computing.*

## 1. INTRODUCTION

Texture as a primitive visual cue has been studied for a long time. Various techniques have been developed for texture segmentation, texture classification and texture synthesis. Although texture analysis has a long history, its applications to real image data have been limited to-date. An important and emerging application where texture analysis can make a significant contribution is the area of content-based retrieval in large image and video databases. Using texture as a visual feature, one can query a database to retrieve similar patterns. Texture classification and segmentation schemes are very important in answering such queries.

Statistical approaches are used to extract texture features. For the analysis of a texture image, it requires large storage space and a lot of computation time to calculate the matrix of features such as SGLDM (Spatial Gray Level Dependence Matrix), NGLDM (Neighboring Gray Level Dependence Matrix) [1] etc. In spite of the large size of each matrix, a set of their scalar feature calculated from the matrix is not efficient to represent the characteristics of image content.

In general, neighboring pixels within an image tend to be highly correlated. As such, it is desired to use an invertible transform to concentrate randomness into fewer, decorrelated parameters. The Discrete Cosine Transform (DCT) has been shown to be near optimal for a large class of images in energy concentration and decorrelating. It has been adopted in the JPEG and MPEG coding standards [2][3]. The DCT decomposes the signal into underlying spatial frequencies, which then allow further processing techniques to reduce the precision of the DCT coefficients consistent with the Human Visual System (HVS) model. The DCT coefficients of an image tend themselves as a new feature, which have the ability to represent the regularity, complexity and some texture features of an image and it can be directly applied to image data in the compressed domain [4]. This may be a way to solve the large storage space problem and the computational complexity of the existing methods.

Soft computing was first proposed by Zadeh [5] to construct new generation computationally intelligent hybrid systems consisting of neural networks, fuzzy inference system, approximate reasoning and derivative free optimization techniques. It is well known that the intelligent systems, which can provide human like expertise such as domain knowledge, uncertain reasoning, and adaptation to a noisy and time varying environment, are important in tackling practical computing problems. In contrast with conventional artificial intelligence techniques which only deal with precision, certainty and rigor the guiding principle of hybrid systems is to exploit the tolerance for imprecision, uncertainty, low solution cost, robustness, partial truth to achieve tractability, and better rapport with reality.

In our research, we used an artificial neural network trained using backpropagation algorithm and an evolving fuzzy neural network (neuro-fuzzy system) [6] for classifying the texture data [7]. The soft computing models were evaluated based on their classification efficiency of the different texture data sets. We also evaluated the performance of the neural network by increasing the training epochs. Some theoretical background about DCT transform is presented in Section 2 followed by texture feature extraction in Section 3. Section 4 and 5 present some basic aspects of neural networks and neuro-fuzzy systems followed by experiment set up in Section 6. Some discussions and conclusions are provided towards the end.

## 2. BLOCK DCT-BASED TRANSFORM

Most existing approaches in texture feature extraction use statistical methods. For the analysis of a texture image, it requires large storage space and a lot of computational time to calculate the matrix of features. For solving the problems, some researchers proposed to use DCT [4] for texture representation. The block diagram of the proposed DCT model is shown in Fig. 1.

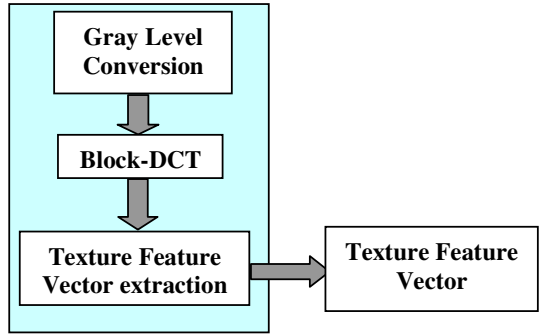

Fig. 1. The block diagram of the texture feature extraction method.

For the DCT transform, we convert an RGB image into gray level image. For spatial localization, we then use the block-based DCT transformation. Each image is divided into $N*N$ sized sub-blocks. The two dimensional DCT can be written in terms of pixel values $f(i, j)$ for $i,j = 0,1,…,N-1$ and the frequency-domain transform coefficients $F(u,v)$:

$$F(u,v) = \frac{1}{\sqrt{2N}} C(u)C(v) \sum_{i=0}^{N-1} \sum_{j=0}^{N-1} f(i, j)$$

$$\times \cos\left[\frac{(2i+1)u\pi}{2N}\right] \cdot \cos\left[\frac{(2j+1)v\pi}{2N}\right] \qquad (1)$$

for $u,v = 0,1,\ldots,N-1$

where

$$c(x) = \begin{cases} \dfrac{1}{\sqrt{2}} & \text{for } x=0 \\ 1 & \text{otherwise} \end{cases}$$

The inverse DCT transform is given by

$$f(i,j) = \sum_{u=0}^{N-1}\sum_{v=0}^{N-1} c(u)c(v)F(u,v)$$

$$\times \cos\left[\frac{(2i+1)u\pi}{2N}\right] \cdot \cos\left[\frac{(2j+1)v\pi}{2N}\right] \quad (2)$$

for $i,j = 0.1,\ldots, N-1$.

## 3. TEXTURE FEATURE VECTOR EXTRACTION

For efficient texture feature extraction, we use some DCT coefficients in compressed domain as the feature vectors. Each sub block contains one DC coefficients and other AC coefficients as shown in Figure 2.

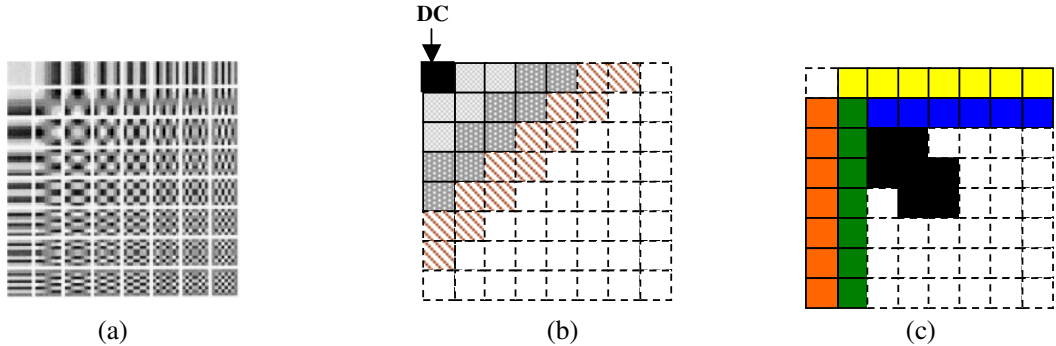

Fig 2. (a) DCT Basis Pattern    (b) Vector element from frequency components   (c) Vector elements from directional information.

Since it is well known that the HVS is less sensitive to errors for high frequency coefficients than it is for lower frequencies component of DCT, we extract the feature set of 9 vector components in which first one is DC coefficient of each sub-block which represents the average energy or intensity of the block and other 8 AC coefficients which represent some different pattern of image variation and directional information of the texture; for example, the coefficients of the most upper region and those of the most left region in a DCT transform domain represent some vertical and horizontal edge information, respectively in Fig. 2(c ).

## 4. ARTIFICIAL NEURAL NETWORK (ANN)

Neural networks are computer algorithms inspired by the way information is processed in the nervous system [8]. An important difference between neural networks and other AI techniques is their ability to learn. The network "learns" by adjusting the interconnections (called weights) between layers. When the network is adequately trained, it is able to generalize relevant output for a set of input data. A valuable property of neural networks is that of generalisation, whereby a trained neural network is able to provide a correct matching in the form of output data for a set of previously unseen input data. Learning typically occurs by example through training, where the training algorithm iteratively adjusts the connection weights (synapses). Backpropagation (BP) is one of the most famous training algorithms for multilayer perceptrons. BP is a gradient descent technique to minimize the error $E$ for a particular training pattern. For adjusting the weight $w_{ij}$ from the $i$-th input unit to the $j^{th}$ output, in the batched mode variant the descent is based on the gradient $\nabla E$ ($\frac{\delta E}{\delta w_{ij}}$) for the total training set:

$$\Delta w_{ij}(n) = -\varepsilon^* \frac{\delta E}{\delta w_{ij}} + \alpha^* \Delta w_{ij}(n-1) \qquad (3)$$

The gradient gives the direction of error $E$. The parameters $\varepsilon$ and $\alpha$ are the learning rate and momentum respectively [9].

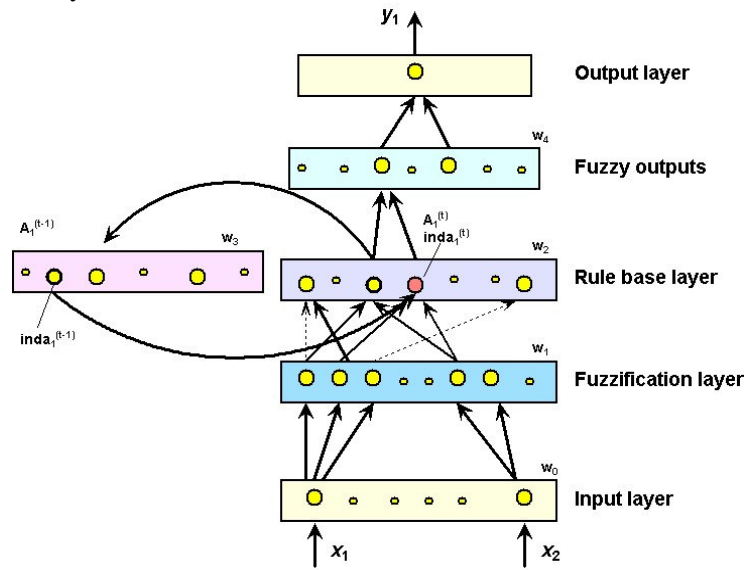

Fig. 3. Architecture of EFuNN.

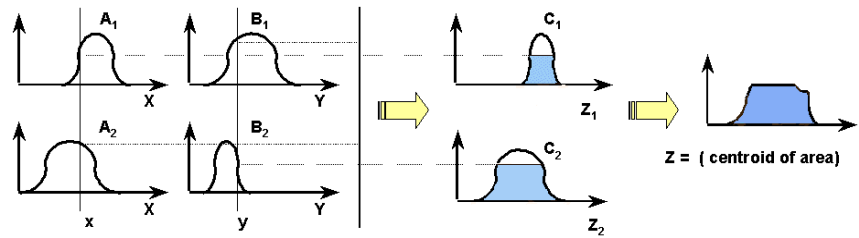

Fig. 4. Mamdani fuzzy inference system.

## 5. NEURO-FUZZY SYSTEMS

A neuro-fuzzy system [10] is defined as a combination of ANN and Fuzzy Inference System (FIS) [11] in such a way that neural network learning algorithms are used to determine the parameters of FIS. An even more important aspect is that the system should always be interpretable in terms of fuzzy *if-then* rules, because it is based on the fuzzy system reflecting vague knowledge. We used Evolving Fuzzy Neural Network (EFuNN) implementing a Mamdani [12] type FIS and all nodes are created during learning. EFuNN has a five-layer structure as illustrated in Figure 3. Fig. 4 illustrates a Mamdani FIS combining 2 fuzzy rules using the *max-min* method [12]. According to the Mamdani FIS, the rule antecedents and consequents are defined by fuzzy sets and has the following structure:

$$\text{if } x \text{ is } A_1 \text{ and } y \text{ is } B_1 \text{ then } z_1 = C_1 \tag{4}$$

where $A_1$ and $B_1$ are the fuzzy sets representing input variables and $C_1$ is the fuzzy set representing the output fuzzy set. In EFuNN, the input layer is followed by the second layer of nodes representing fuzzy quantification of each input variable space. Each input variable is represented here by a group of spatially arranged neurons to represent a fuzzy quantization of this variable. Different membership functions (MF) can be attached to these neurons (triangular, Gaussian, etc.). The nodes representing membership functions can be modified during learning. New neurons can evolve in this layer if, for a given input vector, the corresponding variable value does not belong to any of the existing MF to a degree greater than a membership threshold. The third layer contains rule nodes that evolve through hybrid supervised/unsupervised learning. The rule nodes represent prototypes of input-output data associations, graphically represented as an association of hyper-spheres from the fuzzy input and fuzzy output spaces. Each rule node, e.g. $r_j$, represents an association between a hyper-sphere from the fuzzy input space and a hyper-sphere from the fuzzy output space; $W_1(r_j)$ connection weights representing the co-ordinates of the center of the sphere in the fuzzy input space, and $W_2(r_j)$ – the co-ordinates in the fuzzy output space. The radius of an input hyper-sphere of a rule node is defined as (1- *Sthr*), where *Sthr* is the sensitivity threshold parameter defining the minimum activation of a rule node (e.g., $r_1$, previously evolved to represent a data point $(X_{d1}, Y_{d1})$) to an input vector (e.g., $(X_{d2}, Y_{d2})$) in order for the new input vector to be associated with this rule node. Two pairs of fuzzy input-output data vectors $d_1=(X_{d1}, Y_{d1})$ and $d_2=(X_{d2}, Y_{d2})$ will be allocated to the first rule node $r_1$ if they fall into the $r_1$ input sphere and in the $r_1$ output sphere, i.e. the local normalised fuzzy difference between $X_{d1}$ and $X_{d2}$ is smaller than the radius $r$ and the local normalised fuzzy difference between $Y_{d1}$ and $Y_{d2}$ is smaller than an error threshold *Errthr*. The local normalised fuzzy difference between two fuzzy membership vectors $d_{1f}$ and $d_{2f}$ that represent the membership degrees to which two real values $d_1$ and $d_2$ data belong to the pre-defined MF, are calculated as

$$D(d_{1f}, d_{2f}) = sum(abs(d_{1f} - d_{2f}))/sum(d_{1f} + d_{2f}) \tag{5}$$

If data example $d_1 = (X_{d1}, Y_{d1})$, where $X_{d1}$ and $X_{d2}$ are correspondingly the input and the output fuzzy membership degree vectors, and the data example is associated with a rule node $r_1$ with a centre $r_1^1$, then a new data point $d_2=(X_{d2}, Y_{d2})$, will also be associated with this rule node through the process of associating (learning) new data points to a rule node [6]. The centres of this node hyper-spheres adjust in the fuzzy input space depending on a learning rate $lr_1$, and in the fuzzy output space depending on a learning rate $lr_2$, on the two data point's $d_1$ and $d_2$. The adjustment of the centre $r_1^1$ to its new position $r_1^2$ can be represented mathematically by the change in the connection weights of the rule node $r_1$ from $W_1(r_1^1)$ and $W_2(r_1^1)$ to $W_1(r_1^2)$ and $W_2(r_1^2)$ according to the following vector operations:

$$W_2(r_1^2) = W_2(r_1^1) + lr_2 \cdot Err(Y_{d1}, Y_{d2}) \cdot A_1(r_1^1) \tag{6}$$

$$W_1(r_1^2) = W_1(r_1^1) + lr_1 \cdot Ds(X_{d1}, X_{d2}) \tag{7}$$

where $Err(Y_{d1}, Y_{d2}) = Ds(Y_{d1}, Y_{d2}) = Y_{d1} - Y_{d2}$ is the signed value rather than the absolute value of the fuzzy difference vector; $A_1(r_1^1)$ is the activation of the rule node $r_1^1$ for the input vector $X_{d2}$.

While the connection weights from $W_1$ and $W_2$ capture spatial characteristics of the learned data (centres of hyper-spheres), the temporal layer of connection weights $W_3$ captures temporal dependencies between consecutive data examples. If the winning rule node at the moment $(t-1)$ (to which the input data vector at the moment $(t-1)$ was associated) was $r_1 = inda_1(t-1)$, and the winning node at the moment $t$ is $r_2 = inda_1(t)$, then a link between the two nodes is established as follows:

$$W_3(r_1, r_2)^{(t)} = W_3(r_1, r_2)^{(t-1)} + lr_3 \cdot A_1(r_1)^{(t-1)} A_1(r_2))^{(t)}, \tag{8}$$

where: $A_1(r)^{(t)}$ denotes the activation of a rule node $r$ at a time moment $(t)$; $lr_3$ defines the degree to which the EFuNN associates links between rules (clusters, prototypes) that include consecutive data examples (if $lr_3 = 0$, no temporal associations are learned in an EFuNN structure) [6].

The learned temporal associations can be used to support the activation of rule nodes based on temporal, pattern similarity. Here, temporal dependencies are learned through establishing structural links. The ratio spatial-similarity/temporal-correlation can be balanced for different applications through two parameters $S_s$ and $T_c$ such that the activation of a rule node $r$ for a new data example $d_{new}$ is defined as the following vector operations:

$$A_1(r) = f(S_s \cdot D(r, d_{new}) + T_c \cdot W_3(r^{(t-1)}, r)) \tag{9}$$

where $f$ is the activation function of the rule node $r$, $D(r, d_{new})$ is the normalised fuzzy distance value and $r^{(t-1)}$ is the winning neuron at the previous time moment.

The fourth layer of neurons represents fuzzy quantification for the output variables. The fifth layer represents the real values for the output variables.

EFuNN evolving algorithm is adapted from [6] and is formulated as follows:

1. Initialize an EFuNN structure with a maximum number of neurons and zero value connections. If initially there are no rule nodes connected to the fuzzy input and fuzzy output neurons, then create the first node $r_j = 1$ to represent the first data example $EX = (X_{d1}, Y_{d1})$ and set its input $W_1(r_j)$ and output $W_2(r_j)$ connection weights as follows:
   <Create a new rule node $r_j$> to represent a data sample $EX$: $W_1(r_j) = EX$: $W_2(r_j) = TE$, where $TE$ is the fuzzy output vector for the (fuzzy) example $EX$.
2. While <there are data examples> Do
   Enter the current, example $(X_{di}, Y_{di})$, $EX$ being the fuzzy input vector (the vector of the degrees to which the input values belong to the input membership functions). If there are new variables that appear in this example and have not been used in previous examples, create new input and/or output nodes with their corresponding membership functions.
3. Find the normalized fuzzy similarity between the new example $EX$ (fuzzy input vector) and the already stored patterns in the case nodes $r_j = r_1, r_2, ..., r_n$
   $D(EX, r_j) = sum(abs(EX - W_1(r_j))) / sum(W_1(r_j) + EX)$

4. Find the activation $A_1(r_j)$ of the rule nodes $r_j = r_1, r_2, ..., r_n$. Here radial basis activation (*radbas*) function, or a saturated linear (*satlin*) one, can be used, i.e.
   $A_1(r_j) = radbas(S_s D(EX, r_j - T_c W_3))$, or $A_1(r_j) = satlin(1 - S_s D(EX, r_j + T_c W_3))$.
5. Update the pruning parameter values for the rule nodes, e.g. age, average activation as pre-defined.
6. Find $m$ case nodes $r_j$ with an activation value $A_1(r_j)$ above a predefined sensitivity threshold *Sthr*.
7. From the $m$ case nodes, find one rule node $inda_1$ that has the maximum activation value $maxa_1$.
8. If $maxa_1 < Sthr$, then, <create a new rule node> using the procedure from step 1.
   Else
9. Propagate the activation of the chosen set of $m$ rule nodes $(r_{j1},...,r_{jm})$ to the fuzzy output neurons: $A_2 = satlin(A_1(r_{j1},...,r_{jm}) \cdot W_2)$
10. Calculate the fuzzy output error vector: $Err = A_2 - TE$
11. If $(D(A_2, TE) > Errthr)$ <create a new rule node> using the procedure from step 1.
12. Update (a) the input, and (b) the output of the $m-1$ rule nodes $k = 2 : j_m$ in case of a new node was created, or $m$ rule nodes $k = j_1 : j_m$, in case of no new rule was created:
    - $Ds(EX - W_1(r_k)) = EX - W_1(r_k)$; $W_1(r_k) = W_1(r_k) + lr_1 \cdot Ds(EX - W_1(r_k))$, where $lr_1$ is the learning rate for the first layer;
    - $A_2(r_k) = satlin(W_2(r_k) \cdot A_1(r_k))$; $Err(rk) = TE - A_2(r_k)$;
    - $W_2(r_k) = W_2(r_k) + lr_2 \cdot Err(r_k) \cdot A_1(r_k)$, where $lr_2$ is the learning rate for the second layer.
13. Prune rule nodes $r_j$ and their connections that satisfy the following fuzzy pruning rule to a pre-defined level representing the current need of pruning:
    IF (a rule node $r_j$ is OLD) and (average activation $A_1av(r_j)$ is LOW) and (the density of the neighboring area of neurons is HIGH or MODERATE) (i.e. there are other prototypical nodes that overlap with j in the input-output space; this condition apply only for some strategies of inserting rule nodes as explained below) THEN the probability of pruning node $(r_j)$ is HIGH. The above pruning rule is fuzzy and it requires that the fuzzy concepts as OLD, HIGH, etc. are predefined.
14. Aggregate rule nodes, if necessary, into a smaller number of nodes. A C-means clustering algorithm can be used for this purpose.
15. End of the *while* loop and the algorithm

The rules that represent the rule nodes need to be aggregated in clusters of rules. The degree of aggregation can vary depending on the level of granularity needed. At any time (phase) of the evolving (learning) process, fuzzy, or exact rules can be inserted and extracted [13]. Insertion of fuzzy rules is achieved through setting a new rule node for each new rule, such as the connection weights $W_1$ and $W_2$ of the rule node represent the fuzzy or the exact rule. The process of rule extraction can be performed as aggregation of several rule nodes into larger hyper-spheres. For the aggregation of two-rule nodes $r_1$ and $r_2$, the following aggregation rule is used:

$$If (D(W_1(r_1), W_1(r_2)) <= Thr_1) \text{ and } (D(W_2(r_1), W_2(r_2)) <= Thr_2) \tag{10}$$

then aggregate $r_1$ and $r_2$ into $r_{agg}$ and calculate the centres of the new rule node as:

$$W_1(r_{agg}) = average(W_1(r_1), W_1(r_2)), W_2(r_{agg}) = average(W_2(r_1), W_2(r_2)) \tag{11}$$

Here the geometrical center between two points in a fuzzy problem space is calculated with the use of an average vector operation over the two fuzzy vectors. This is based on a presumed piece-wise linear function between two points from the defined through the parameters *Sthr* and *Errthr* input and output fuzzy hyper-spheres.

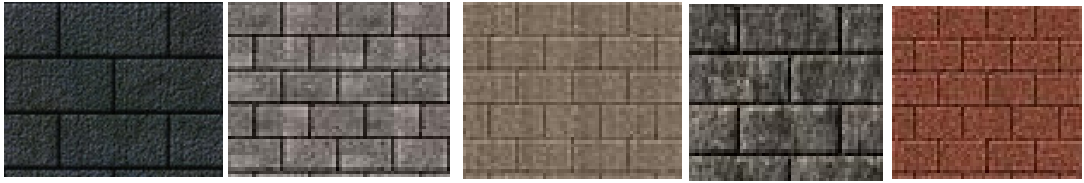

(a) Brick

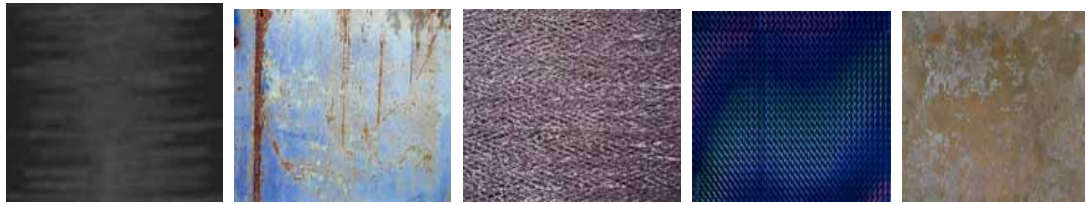

(b) Metal

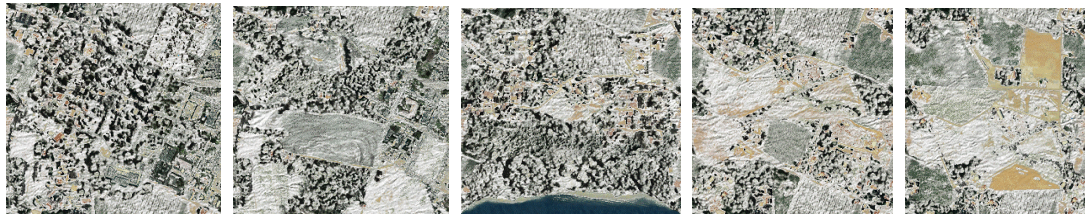

(c) Rural

Fig. 5. Some sample texture patterns from the database.

## 6. EXPERIMENT SETUP AND RESULTS

### 6.1 Training and Testing Data

In our research, we attempted to classify 3 different types of textures using soft computing techniques. We used DCT coefficients to represent the different textures. Each texture image was represented by 324 DCT coefficients. Our texture database consisted of 240 different textures and we manually classified into three different classes (*brick*, *metal* and *rural*). Some sample textures are illustrated in Figure 5. 192 texture datasets were used for training the soft computing models and remaining 48 texture datasets were used for testing purposes.

While the proposed neuro-fuzzy model was capable of determining the architecture automatically, we had to do some initial experiments to determine the architecture (number of hidden neurons and number of layers) of the neural network. After a trial and error approach we found that the neural network was giving good generalization performance when we had 2 hidden layers with 90 neurons each. In the following sections we report the details of our experimentations with neural networks and neuro-fuzzy models.

### 6.2 EFuNN Training

We used 4 membership functions for each input variable and the following evolving parameters: sensitivity threshold *Sthr*=0.99, error threshold *Errthr*=0.001. EFuNN training has created 162 rule nodes as shown in Figure 6. Empirical results are reported in Tables 1 and 2.

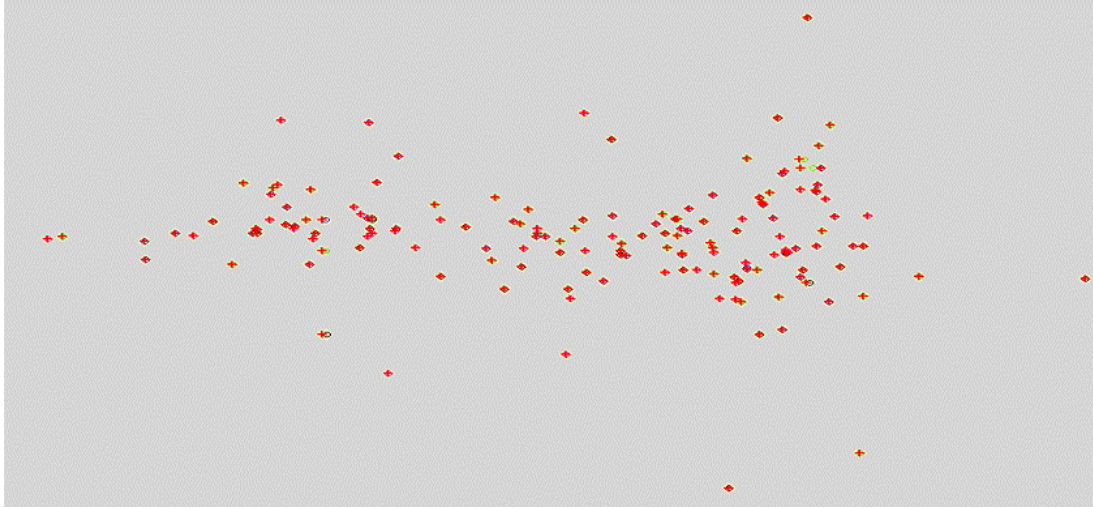

Fig. 6. Learned rule nodes by EFuNN learning.

### 6.3 ANN Training

We used a neural network trained using backpropagation algorithm. We used 1 input layer, 2 hidden layers and an output layer [327-90-90-3]. Input layer consists of 327 neurons corresponding to the input variables. The first and second hidden layer consists of 90 neurons each with tanh-sigmoidal activation function. The initial learning rate and momentum were set as 0.05 and 0.1 respectively. Training errors (RMSE) and performance achieved after 5000, 15,000, 20,000 and 40,000 epochs are reported in Tables 1 and 2. Approximate computational load in Giga Flops is reported in Table 1 and is graphically depicted in Figure 7.

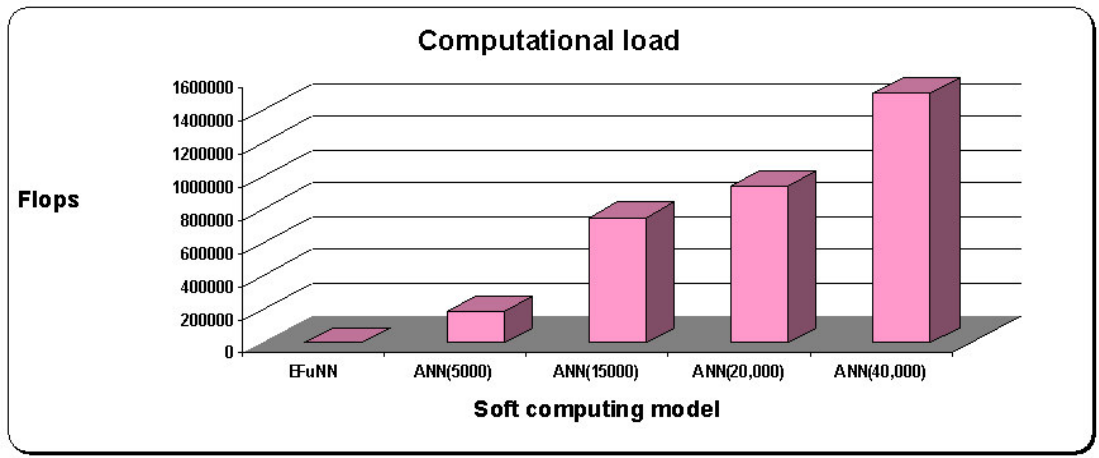

Fig. 7. Computational load of soft computing models.

### 6.4 Test Results

Table 2 summarizes the comparative performance of EFuNN and ANN. The best classification was obtained using EFuNN (88%) and 85.4% for neural networks.

Table 1. Training performance of Soft computing models

|  | EFuNN | ANN (5000 ep) | ANN (15000 ep) | ANN (20000 ep) | ANN (40,000 ep) |
|---|---|---|---|---|---|
| RMSE (Training) | $0.2e^{-003}$ | $3.9e^{-003}$ | $1.4e^{-004}$ | $9.4e^{-005}$ | $8.5e^{-005}$ |
| Giga Flops | 152 | 187860 | 751100 | 939000 | 1502200 |

Table 2. Test results and performance comparison of texture classification

|  |  | EFuNN | ANN (5000 ep) | ANN (15000 ep) | ANN (20000 ep) | ANN (40,000 ep) |
|---|---|---|---|---|---|---|
| *Brick* (16 Nos) | $A_1$ | 15 | 14 | 14 | 12 | 12 |
|  | $A_2$ | 1 | 2 | 2 | 4 | 4 |
| *Metal* (16 Nos) | $B_1$ | 16 | 13 | 14 | 15 | 15 |
|  | $B_2$ | 0 | 3 | 2 | 1 | 1 |
| *Rural* (16 Nos) | $C_1$ | 11 | 12 | 11 | 14 |  |
|  | $C_2$ | 5 | 4 | 5 | 2 | 2 |
| Total (48 Nos) | $X = (A_1+B_1+C_1)$ | 42 | 39 | 39 | 41 | 41 |
|  | $Y = (A_2+B_2+C_2)$ | 6 | 9 | 9 | 7 | 7 |
| *Reliability of classification |  | 88 % | 81.25 % | 81.25 % | 85.4 % | 85.4 % |

*Reliability = $\left(\frac{X}{48}\right) \times 100$

## 7. CONCLUSIONS

In this paper, we attempted to classify 3 different types of textures using artificial neural networks and Evolving Fuzzy Neural Network (EFuNN). For texture feature we considered the DCT coefficient, which does not require additional complex computation for feature extraction. As the high frequency coefficient is less sensitive to human visual systems, we constructed a feature matrix consisting of the first few coefficients of each block. EFuNN outperformed the neural network with the best classification (88%). As depicted in Table 1, EFuNN is less computational expensive while compared to neural networks. EFuNN adopts a one-pass (one epoch) training technique, which is highly suitable for online learning. Hence online training can incorporate further knowledge very easily. We also studied the generalization performance of BP training when the training epochs were increased from 5000 epochs to 40,000 epochs. When the number of epochs were increased, it was interesting to note the continuous reduction of the training error (RMSE) but the generalization error (classification accuracy) however tends to settle after 20,000 epochs. Compared to ANN, an important advantage of neuro-fuzzy model is its reasoning ability (*if-then* rules) of any particular state [13].

The proposed prediction models based on soft computing on the other hand are easy to implement and produces desirable mapping function by training on the given data set. Moreover, the considered connectionist models are very robust, capable of handling the noisy and approximate data and might be more reliable in worst situations. Choosing suitable parameters for the soft computing models is more or less a trial and error approach. Optimal results will depend on the selection of parameters. Selection of optimal parameters may be formulated as an evolutionary search to make SC models fully adaptable and optimal according to the requirement [9].

## REFERENCES

[1]     F. Borko, Video and Image processing in multimedia Systems, Kluwer Academic publishers, 1995, pp 225-249.
[2]     G. K. Wallace, " Overview of the JPEG still Image Compression standard," SPIE 1244 (1990) 220-233.
[3]     D. J. Le Gall, " The MPEG Video Compression Algorithm: A review," SPIE 1452 (1991) 444-457.
[4]     Sang-Mi Lee, Hee_Jung Bae, and Sung-Hwan Jung, "Efficient Content-Based Image Retrieval Methods Using Color and Texture", *ETRI Journal* 20 (1998) 272-283.
[5]     LA. Zadeh, Roles of Soft Computing and Fuzzy Logic in the Conception, Design and Deployment of Information/Intelligent Systems, Computational Intelligence: Soft Computing and Fuzzy-Neuro Integration with Applications, O Kaynak, LA Zadeh, B Turksen, IJ Rudas (Eds.), pp1-9, 1998.
[6]     N. Kasabov, "Evolving Fuzzy Neural Networks - Algorithms, Applications and Biological Motivation, in Yamakawa T and Matsumoto G (Eds), Methodologies for the Conception, Design and Application of Soft Computing", *World Scientific*, pp. 271-274, 1998.
[7]     G.Sorwar, A. Abraham and L. Dooley, "Texture Classification Based on DCT and Soft Computing", in *the 10th IEEE International Conference on Fuzzy Systems*, December 2001, Melbourne, Australia.
[8]     J. M. Zurada, Introduction to Artificial Neural Systems, PWS Pub Co, 1992.
[9]     A. Abraham, Meta-Learning Evolutionary Artificial Neural Networks, Neurocomputing Journal, Elsevier Science, Netherlands, Vol. 56c, pp. 1-38, 2004.
[10]    A. Abraham, Neuro-Fuzzy Systems: State-of-the-Art Modeling Techniques, Connectionist Models of Neurons, Learning Processes, and Artificial Intelligence, Lecture Notes in Computer Science. Volume. 2084, Springer Verlag Germany, Jose Mira and Alberto Prieto (Eds.), ISBN 3540422358, Spain, pp. 269-276, 2001.
[11]    V. Cherkassky, "Fuzzy Inference Systems: A Critical Review, Computational Intelligence: Soft Computing and Fuzzy-Neuro Integration with Applications", Kayak O, Zadeh LA et al (Eds.), Springer, pp.177-197, 1998.
[12]    E. M. Mamdani and S. Assilian, "An experiment in Linguistic Synthesis with a Fuzzy Logic Controller", *International Journal of Man-Machine Studies*, **7**(1), pp. 1-13, 1975.